\pdfoutput=1

\documentclass{article}

\usepackage[nonatbib,preprint]{neurips_2024}
\usepackage{graphicx}
\usepackage[utf8]{inputenc} % allow utf-8 input
\usepackage[T1]{fontenc}    % use 8-bit T1 fonts
\usepackage{hyperref}       % hyperlinks
\usepackage{url}            % simple URL typesetting
\usepackage{amsfonts}       % blackboard math symbols
\usepackage{nicefrac}       % compact symbols for 1/2, etc.
\usepackage{microtype}      % microtypography
\usepackage{xcolor}         % colors
\usepackage{inconsolata}

\usepackage{amsmath}
\usepackage{threeparttable}

\usepackage[capitalize,noabbrev]{cleveref}
\crefformat{section}{\S#2#1#3}
\Crefformat{section}{\S#2#1#3}
\crefmultiformat{section}{\S#2#1#3}{ and \S#2#1#3}{, \S#2#1#3}{ and \S#2#1#3}
\Crefmultiformat{section}{\S#2#1#3}{ and \S#2#1#3}{, \S#2#1#3}{ and \S#2#1#3}

\usepackage{tabularray}
\UseTblrLibrary{booktabs}
\SetTblrInner[booktabs]{abovesep=0pt, belowsep=0pt, rowsep=0.15pt, colsep=3.5pt}
\SetTblrInner[booktabs]{cells = {cmd=\small}}
\NewTableCommand\seprule{\specialrule{\lightrulewidth,gray8}{2pt}{2pt}}
\NewTableCommand\uniquerule{\specialrule{\lightrulewidth,gray7,dashed}{2pt}{2pt}}
\definecolor{lightb}{RGB}{235,245,255}

\newcommand{\snowflake}[1]{{#1\textsuperscript{\textnormal{1}}}}
\newcommand{\snowflakefront}[1]{{\textsuperscript{1}#1}}
\newcommand{\uiuc}[1]{{#1\textsuperscript{\textnormal{2}}}}
\newcommand{\uiucfront}[1]{{\textsuperscript{2}#1}}

\newcommand{\snufront}[1]{{\textsuperscript{3}#1}}

\newcommand{\snusnowflake}[1]{{#1\textsuperscript{\textnormal{1,3}}}}

\usepackage{xspace}
\newcommand{\ours}{\mbox{Arctic-SnowCoder}\xspace}
\newcommand{\oursone}{\mbox{Arctic-SnowCoder-alpha}\xspace}
\newcommand{\ourstwo}{\mbox{Arctic-SnowCoder-beta}\xspace}
\newcommand{\llm}{LLM\xspace}

\newcommand{\stack}{The Stack\xspace}
\newcommand\finewebedu{FineWeb-Edu\xspace}
\newcommand\fineweb{FineWeb\xspace}
\newcommand\dclm{DCLM\xspace}
\newcommand\github{GitHub\xspace}
\newcommand\fasttext{\texttt{fastText}\xspace}

\newcommand\humaneval{HumanEval\xspace}
\newcommand\humanevalp{HumanEval+\xspace}
\newcommand\evalplus{EvalPlus\xspace}
\newcommand\mbpp{MBPP\xspace}
\newcommand\mbppp{MBPP+\xspace}
\newcommand\evoeval{EvoEval\xspace}
\newcommand\bigcodebench{BigCodeBench\xspace}

\newcommand\deepseektwo{DeepSeek-Coder-V2\xspace}
\newcommand\deepseekmath{DeepSeek-Math\xspace}
\newcommand\llamatwo{Llama-2\xspace}
\newcommand\llamathree{Llama-3\xspace}
\newcommand\llamathreeone{Llama-3.1\xspace}

\newcommand\starcoderbase{StarCoderBase\xspace}
\newcommand\starcodertwo{StarCoder2\xspace}
\newcommand\stablecode{StableCode\xspace}
\newcommand\deepseek{DeepSeek-Coder\xspace}
\newcommand\codegemma{CodeGemma\xspace}
\newcommand\granit{Granite-Code\xspace}
\newcommand\codegen{CodeGen\xspace}

\newcommand\magicoder{Magicoder\xspace}
\newcommand\ossinstruct{OSS-Instruct\xspace}

\newcommand\scins{StarCoder2-Instruct\xspace}
\newcommand\smol{SmolLM\xspace}
\newcommand\qwen{Qwen\xspace}
\newcommand\phionefive{Phi-1.5\xspace}
\newcommand\phims{Phi\xspace}
\newcommand\snowarcm{\texttt{Snowflake-arctic-embed-m}}
\newcommand\mixtral{Mixtral}

\newcommand\annEdu{\textsc{Ann-Edu}\xspace}
\newcommand\annIns{\textsc{Ann-Ins}\xspace}
\newcommand\annHq{\textsc{Ann-HQ}\xspace}
\newcommand\annBest{\textsc{Ann-HQIns}\xspace}

\newcommand\mathMaxLR{\texttt{MAX\_LR}}
\newcommand\mathMinLR{\texttt{MIN\_LR}}
\usepackage{comment}
\usepackage{pifont}
\usepackage{tabularx}
\usepackage{caption, booktabs}
\usepackage{algorithm}
\usepackage{algpseudocode}
\usepackage{amssymb}
\usepackage{setspace}
\usepackage{bbm}
\usepackage{arydshln}

\usepackage{graphicx}
\usepackage{hyperref}
\hypersetup{
    colorlinks=true,
    linkcolor=violet,
    citecolor=violet,
    urlcolor=blue
}

\title{\ours: Demystifying High-Quality Data in Code Pretraining}

\newcommand{\separate}{{\quad}}
\author{
  \uiuc{Yuxiang Wei}\thanks{Work done during an internship at Snowflake AI Research.}\separate
  \snusnowflake{Hojae Han}\separate
  \snowflake{Rajhans Samdani}\separate
  % ...
  \\[\smallskipamount]
  \snowflakefront{Snowflake AI Research}
  \\[\smallskipamount]
  \uiucfront{University of Illinois at Urbana-Champaign}
  \separate
  \snufront{Seoul National University}
      \\[\smallskipamount]
    {\normalsize \texttt{\texttt{ywei40@illinois.edu}\ \ \{hojae.han,rajhans.samdani\}@snowflake.com}}
}

\begin{document}
\maketitle

\begin{abstract}
Recent studies have been increasingly demonstrating that high-quality data is crucial for effective pretraining of language models. However, the precise definition of ``high-quality'' remains underexplored. Focusing on the code domain, we introduce \ours-1.3B, a data-efficient base code model pretrained on 555B tokens through three phases of progressively refined data:
(1) \emph{general pretraining} with 500B standard-quality code tokens, preprocessed through basic filtering, deduplication, and decontamination,
(2) \emph{continued pretraining} with 50B high-quality tokens, selected from phase one by a BERT-style quality annotator trained to distinguish good code from random data, using positive examples drawn from high-quality code files, along with instruction data from \magicoder and \scins,
and (3) \emph{enhanced pretraining} with 5B synthetic data created by \llamathreeone-70B using phase two data as seeds, adapting the \magicoder approach for pretraining. 
Despite being trained on a limited dataset, \ours achieves state-of-the-art performance on \bigcodebench{}, a coding benchmark focusing on practical and challenging programming tasks, compared to similarly sized models trained on no more than 1T tokens, outperforming \phims-1.5-1.3B by 36\%.
Across all evaluated benchmarks, \ours-1.3B beats \starcoderbase-3B pretrained on 1T tokens.
Additionally, it matches the performance of leading small base code models trained on trillions of tokens.
For example, \ours-1.3B surpasses \starcodertwo-3B, pretrained on over 3.3T tokens, on \humanevalp, a benchmark that evaluates function-level code generation, and remains competitive on \bigcodebench{}. Our evaluation presents a comprehensive analysis justifying various design choices for \ours. Most importantly, we find that the key to high-quality data is its alignment with the distribution of downstream applications.
\end{abstract}

\begin{figure}[htbp]
\centering
\includegraphics[width=\linewidth]{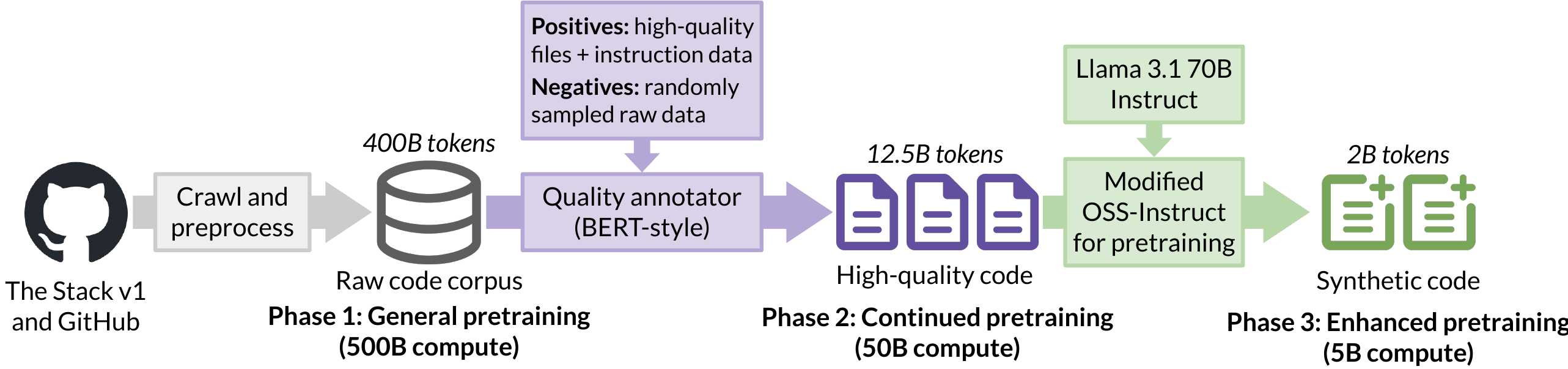}
\caption{Three-phase pretraining of \ours{}-1.3B with progressively higher-quality data.}
\label{fig:overview}
\end{figure}

\section{Introduction}
\label{intro}
Pretraining large language models (LLMs) has generally relied on vast quantities of data. This emphasis on data volume is especially true in specialized domains like code, where researchers obtain massive code pretraining datasets by crawling platforms like \github~\cite{li2023starcoder, codellama, dscoder, starcoder2, granite, dscoderv2}.
Recent studies, however, have increasingly showed that high-quality data is crucial for effective pretraining~\cite{deepseekv2,fineweb,dclm,phi3}, including the code domain~\cite{phi1,phi15,dscoderv2}.
% Despite this recognition, the definition of "high-quality" data remains unclear.

In the general domain, researchers have explored various techniques to curate high-quality pretraining data for language models. \finewebedu~\cite{fineweb} uses a linear regressor built on \snowarcm~\cite{arcticembed} embeddings to assess the educational value of web pages and select high-quality content, while the \dclm~\cite{dclm} approach employs a \fasttext-based~\cite{fasttext} filter trained on positive examples from high-quality online sources~\cite{hqcode} and instruction data~\cite{magicoder}, and random negative web pages to identify high-quality text. These model-based quality filters have been shown to significantly enhance language model performance on downstream tasks, compared to using unfiltered, large-scale datasets. Similarly, researchers have recognized the importance of high-quality code data for pretraining, with \phims-1~\cite{phi1} using a random forest classifier on \codegen~\cite{nijkamp2023codegen} embeddings to select educational code samples, and \deepseektwo~\cite{deepseekv2} employing a multi-stage \fasttext-based~\cite{fasttext} pipeline to recall web-related code data and high-quality code from \github, achieving state-of-the-art coding performance.

% Similarly, \ours utilizes a high-quality code annotator to extract and generate synthetic code files from the pretraining dataset, aiming to further improve the model's coding capabilities by tailoring the data to match the target downstream tasks.

In this paper, we introduce \ours-1.3B, a high-performing small code model created by a novel three-step training methodology focused on progressive improvements in data quality. As a result of this methodology, \ours-1.3B outperforms \starcoderbase-3B~\cite{li2023starcoder} across all evaluated benchmarks and exceeds \phims-1.5-1.3B~\cite{phi15} by 36\% on the complex and practical \bigcodebench benchmark~\cite{bigcodebench}, a benchmark that truly matters for real-world programming.
% In this paper, we introduce \ours-1.3B, a small code model designed to tackle the challenge of defining and leveraging high-quality data in code pretraining.
As shown in \Cref{fig:overview}, \ours is developed through a three-stage, data-efficient pretraining process that progressively refines the quality of the data used. The first stage involves general pretraining for a 500B token horizon using 400B unique raw code data, which have been preprocessed through basic filtering, deduplication, and decontamination. The 400B raw corpus is primarily derived from the coding data used to train Snowflake Arctic~\cite{snowflakearctic},
combining cleaned \stack v1~\cite{li2023starcoder} and \github crawls.
This is followed by continued pretraining on 50B tokens, utilizing a smaller, high-quality subset of 12.5B code files, repeated four times. The high-quality tokens are selected from phase one by a BERT-based~\cite{bert} quality annotator trained to distinguish good code from random data, using positive examples drawn from publicly available high-quality code files~\cite{hqcode}, along with instruction data from \magicoder~\cite{magicoder} and \scins~\cite{sc2instruct}.
Finally, the model undergoes an enhanced pretraining phase for 5B tokens, leveraging roughly 2B synthetic data generated by \llamathreeone-70B~\cite{llama31}. This process uses the phase two data as seeds and adapts the OSS-Instruct methodology from \magicoder~\cite{magicoder} by transforming lower-quality seed code into high-quality code documents.
Notably, all training phases of \ours derive data from the same raw pretraining corpus, ensuring that minimal new knowledge is introduced.

\ours-1.3B achieves state-of-the-art results on \bigcodebench{}~\cite{bigcodebench}, a coding benchmark focusing on practical and challenging programming tasks, among models of similar size trained with $\le$ 1T tokens. Particularly, it outperforming \phims-1.5-1.3B~\cite{phi15} by 36\%. Despite being trained on 555B tokens, compared to other state-of-the-art small code models trained on trillions of tokens, \ours matches or surpasses the performance of these models on several benchmarks. For instance, \ours-1.3B beats \starcoderbase-3B~\cite{li2023starcoder}, trained on over 1T tokens, across all evaluated benchmarks. \ours-1.3B outperforms \starcodertwo-3B~\cite{starcoder2}, trained on over 3T tokens, on \humanevalp~\cite{codex,evalplus} (28.0 vs. 27.4), a benchmark evaluating function-level code generation, while remaining competitive on \bigcodebench (19.4 vs. 21.4).
We conduct comprehensive ablation studies to validate the design decisions behind training \ours:
\begin{itemize}
\item First, our findings indicate that, in general pretraining, organizing file-level data into repositories after partitioning by programming language significantly outperforms the approach of grouping data solely by repository names.
\item Additionally, we determine the optimal learning rate schedule, which involves a re-warmup phase followed by linear decay, as well as the ideal repetition of high-quality data during continued pretraining, which we find to be four times.
\item More importantly, our comparisons of model-based quality annotators, trained on various data combinations, highlight that the alignment of pretraining data with downstream tasks is crucial for achieving superior performance.
\end{itemize}
% Our analysis provides valuable insights into the optimal strategies for data selection and repetition during continued pretraining, offering practical guidelines for enhancing future model development.
% Our analysis offers insights into the optimal design choices for data selection and repetition during continued pretraining, providing practical guidelines for future model development.

In summary, we make the following contributions:

\begin{itemize}
\item We introduce \ours-1.3B, a high-performing small code model trained on 555B tokens that benefits from progressive improvements in data quality.
\item We demonstrate that high-quality data and synthetic data can significantly improve the model performance despite being seeded from the same raw corpus.
\item For the first time, we demystify the notion of data quality in code pretraining by systematically comparing model-based quality annotators trained on different data combinations.
\item We provide practical insights into optimal design choices for repo-level grouping in general pretraining, and optimal learning rate schedules and repetitions of high-quality data during continued pretraining, providing practical guidelines for future model development.
\end{itemize}

\section{\ours}
In this section, we provide a detailed explanation of the training methodology used for \ours-1.3B, as illustrated in \Cref{fig:overview}. We begin by discussing the composition of the raw training data in \Cref{sec:tech:rawdata}, followed by an overview of the general pretraining phase in \Cref{sec:tech:general}. Next, we describe the continued pretraining process using high-quality data in \Cref{sec:tech:continue}, and finally, we elaborate on the enhanced pretraining with synthetic data in \Cref{sec:tech:enhance}. The model architecture is based on \llamatwo~\cite{llama2}, with specific details provided in \Cref{tab:architecture}.

\begin{table*}[htbp]
\caption{Model architecture details of \ours.}
\label{tab:architecture}
\centering
\begin{booktabs}{
    colspec={@{}ll@{}},
    % cell{1}{3}={c=2}{c},
    % cell{1}{1-2}={r=2}{m},
    % % cell{Y}{3-Z}={font=\bfseries}
    % row{X-Z} = {bg=lightb}
}
\toprule
Parameter & \ours-1.3B \\
\midrule
\texttt{hidden\_dim} & 2048 \\
\texttt{ffn\_hidden\_dim} & 5632 \\
\texttt{num\_heads} & 16 \\
\texttt{num\_kv\_heads} & 16 \\
\texttt{num\_layers} & 24 \\
\texttt{vocab\_size} & 64000 \\
\texttt{seq\_len} & 8192 \\
\texttt{positional\_encodings} & RoPE~\cite{rope} \\
\texttt{tie\_embeddings\_and\_output\_weights} & True \\
\bottomrule
\end{booktabs}%
\end{table*}

\subsection{Raw data}
\label{sec:tech:rawdata}
The raw pretraining data used to train \ours-1.3B consists exclusively of code, primarily derived from the coding data used to train Snowflake Arctic~\cite{snowflakearctic}. This data combines cleaned versions of \stack v1~\cite{li2023starcoder} and \github crawls. From this data, we select 18 popular programming languages for training, similar to \starcodertwo-3B~\cite{starcoder2}. These languages include Python, Java, C++, C, JavaScript, PHP, C\#, Go, TypeScript, SQL, Ruby, Rust, Jupyter Notebook, Scala, Kotlin, Shell, Dart, Swift, amounting to a total of 400B unique tokens.

\subsection{General pretraining}
\label{sec:tech:general}
In general pretraining, the model is trained for 500B tokens with a sequence length of 8,192 and a batch size of 512 using Adam~\cite{adam}. The learning rate follows a cosine decay after a linear warmup of 600 iterations. We set the maximum learning rate to $5.3\times 10^{-4}$ and the minimum to $5.3\times10^{-5}$, following \deepseek~\cite{dscoder}.
In this phase, we use the entire 400B raw data without applying additional quality filtering. We start by partitioning code files by programming language, grouping them by repository, and then concatenating them in random order, similar to the \starcodertwo{}~\cite{starcoder2} approach.
In \Cref{sec:ablation:repo}, we show the advantage of first partitioning code files by programming language.
We name the model produced by this phase as \oursone.

\subsection{Continued pretraining with high-quality data}
\label{sec:tech:continue}
After general pretraining, we continue pretraining \oursone with 50B high-quality tokens sourced from the same raw pretraining corpus. The 50B high-quality tokens are formed by repeating 12.5B top-percentile code file tokens for 4 times scored by our code quality annotator.
Inspired by \finewebedu~\cite{fineweb} and \dclm~\cite{dclm},
we train a linear classification head on top of \snowarcm~\cite{arcticembed}, a state-of-the-art embedding model based on BERT~\cite{bert}.
The training data comprises 300k positive examples, sampled from a blend of 220k high-quality open-source code files~\cite{hqcode}, 80k high-quality instruction data from \magicoder~\cite{magicoder} and \scins~\cite{sc2instruct}, and 300 randomly selected code documents from the pretraining corpus.
Prior research on code quality, such as \phims-1~\cite{phi1}, often overemphasizes the ``educational value'' of code, skewing models towards simpler benchmarks like \humanevalp~\cite{codex,evalplus}. In \Cref{sec:eval:all}, we show that our annotation leads to a more balanced enhancement of model capabilities.
Furthermore, given that these code documents typically exceed 1000 tokens, surpassing the BERT context window size of 512, we improve over \finewebedu's pipeline to calculate the score for each file by averaging the scores from the top, middle, and bottom sections as produced by the quality annotator.
In this phase, we rewarmup the learning rate for 1000 iterations from 0 to $5.3\times 10^{-4}$, the maximum pretraining learning rate, followed by a linear decay to 0. The model produced in this phase is referred to as \ourstwo. In \Cref{sec:ablation:continue}, we validate all of our design choices.

\subsection{Enhanced pretraining with synthetic data}
\label{sec:tech:enhance}
In the enhanced pretraining stage, we generate even higher-quality data than in continued pretraining leveraging \llamathreeone-70B-Instruct~\cite{llama31} and increase the Python mix ratio to approximately 50\% while keeping the proportions of the other languages unchanged.
\phims-1~\cite{phi1} demonstrates that synthetic, textbook-like pretraining data can significantly enhance model performance. However, overemphasis on such data risks skewing the model's distribution, potentially impairing its effectiveness in real-world coding tasks.
For example, we show in \Cref{sec:eval:all} that \phims-1.5 excels in \humanevalp~\cite{codex,evalplus} and \mbppp~\cite{mbpp,evalplus}, which resemble textbook exercises, but performs less effectively on the more complex and practical coding tasks in \bigcodebench~\cite{bigcodebench}.
To address this, we adapt the \ossinstruct method from \magicoder~\cite{magicoder} for pretraining purposes. Originally, \ossinstruct was originally designed to generate realistic instruction-tuning data by prompting a model to create question-answer pairs inspired by open-source code snippets. In contrast, we produce high-quality synthetic pretraining data by using \llamathreeone-70B-Instruct to generate high-quality and problem-solving oriented code files, seeded with code documents scored in the top percentile during the continued pretraining phase.
In \Cref{sec:eval:all}, we demonstrate that each pretraining phase significantly outperforms the previous one, highlighting the effectiveness of progressively enhancing data quality.

\section{Experiments}
In this section, we compare \ours with state-of-the-art small language models and show performance boost over each pretraining stage (\Cref{sec:eval:all}), evaluate two strategies of forming repo-level data in general pretraining (\Cref{sec:ablation:repo}), and perform detailed ablation to justify our design choices in continued pretraining (\Cref{sec:ablation:continue}).

\subsection{Experimental setup}
We consider the following four diverse programming benchmarks to comprehensively evaluate the code generation capability of different code models:
\begin{description}
\item[\humanevalp and \mbppp~\cite{evalplus}.]
\humaneval~\cite{codex} and \mbpp~\cite{mbpp} are the two most widely-used benchmarks for function-level code generation. We adopt their augmented version powered by \evalplus~\cite{evalplus}, with 80×/35× more test cases for rigorous evaluation. \humanevalp and \mbppp include 164 and 378 coding problems, respectively.
\item[\evoeval~\cite{evoeval}] is a program synthesis benchmark suite created by evolving existing benchmarks into different targeted domains. We employ its five default transformation categories, namely \texttt{difficult}, \texttt{creative}, \texttt{subtle}, \texttt{combine} and \texttt{tool\_use}, totaling 500 tasks.
\item[\bigcodebench~\cite{bigcodebench}] evaluates LLMs with practical and challenging programming tasks. It has 1140 programming tasks, where each task in \bigcodebench is created through human-LLM collaboration, where the task quality is ensured by human experts.

\end{description}
We incorporate \humanevalp, \mbppp, \evoeval, and \bigcodebench for baseline comparison in \Cref{sec:eval:all}. For the subsequent ablation studies in \cref{sec:ablation:repo,sec:ablation:continue},
we include the base versions of \humaneval and \mbpp while omitting \bigcodebench for faster evaluation.
Throughout the experiments, we report the pass@1 metric~\cite{codex} using greedy decoding.

\subsection{Baseline comparison and effectiveness of three-stage pretraining}
\label{sec:eval:all}
\begin{table*}[htbp]
\caption{Comparing \ours with state-of-the-art small language models (< 3B), divided by whether training compute > 1T tokens. \oursone and \ourstwo are checkpoints after general pretraining and continued pretraining with high-quality data, respectively. \ours is the final checkpoint after enhanced pretraining with synthetic data.}

\label{tab:benchmark-results}
\centering
\begin{booktabs}{
    colspec={@{}lrrrrr@{}},
    % cell{1}{3}={c=2}{c},
    % cell{1}{1-2}={r=2}{m},
    % % cell{Y}{3-Z}={font=\bfseries}
    % row{X-Z} = {bg=lightb}
}
\toprule
Model & Training compute & \humaneval{+} & \mbpp{+} & \evoeval & \bigcodebench \\

\midrule
\stablecode-3B~\cite{stablecode} & 1.3T & 26.2 & 43.9 & 18.6 & 25.9 \\
\starcodertwo-3B~\cite{starcoder2} & 3.3T to 4.3T & 27.4 & \textbf{49.2} & 19.0 & 21.4 \\
\granit-Base-3B~\cite{granite} & 4.5T & 29.3 & 45.8 & \textbf{19.8} & 20.0 \\
\codegemma-2B-v1.0~\cite{codegemma} & 3T + 1T & 18.3 & 46.3 & 15.4 & 23.9 \\
\codegemma-2B-v1.1~\cite{codegemma} & 3T + 500B & \textbf{32.3} & 48.9 & \textbf{19.8} & \textbf{28.0} \\
\qwen{1.5}-1.8B$^1$~\cite{qwen2} & 3T & 19.5 & 28.3 & 5.0 & 6.3 \\
\qwen{2}-1.5B$^1$~\cite{qwen2} & 7T & 31.1 & 38.4 & 17.2 & 16.5 \\
\deepseek-1.3B~\cite{dscoder} & 2T & 28.7 & 48.1 & 19.2 & 22.2 \\

\midrule
% \codegen-3.7B$^2$ & 300B & 12.8 & - & - & - \\
% \codetf-2B$^2$ & > 100B & 22.0 & 38.1 & - & - \\
\starcoderbase-3B~\cite{li2023starcoder} & 1T & 17.7 & 36.8 & 11.6 & 5.9 \\
\smol-1.7B~\cite{smollm} & 1T & 15.9 & 34.7 & 10.0 & 2.5 \\
\phionefive-1.3B~\cite{phi15} & 150B & \textbf{31.7} & \textbf{43.7} & \textbf{20.6} & \textbf{14.3} \\  %  & 14.8 \\
\seprule
\oursone-1.3B & 500B & 14.0 & 27.8 & 7.4 & 10.3 \\%  19.1 \\
\ourstwo-1.3B & 500B + 50B & 21.3 & 34.7 & 12.8 & 12.3 \\ %& 25.0 \\
\ours-1.3B & 550B + 5B & \textbf{28.0} & \textbf{42.9} & \textbf{18.0} & \textbf{19.4} \\% & \textbf{23.0} \\
\bottomrule
\end{booktabs}%

\begin{tablenotes}
\item[*] {\scriptsize \({}^{1}\) We remove trailing newlines from prompts in most \humaneval~(+) and \evoeval evaluations. However, for \qwen{1.5}-1.8B and \qwen{2}-1.5B, we keep them due to their high sensitivity (>15 points drop) to newlines.}
% \item[*] {\scriptsize \({}^{2}\) For \codegen and \codetf, results are collected from the \evalplus leaderboard as they are unsupported by vLLM used in our evaluation.}
\end{tablenotes}
\end{table*}

\Cref{tab:benchmark-results} presents a comprehensive comparison of various small language models (less than 3B parameters) across multiple coding benchmarks, categorized by whether their training compute exceeds 1T tokens. Notably, \ours demonstrates exceptional performance, particularly given its limited training data.
\ours-1.3B achieves state-of-the-art performance on \bigcodebench{}
compared to similarly sized models trained on no more than 1T token, significantly outperforming \starcoderbase-3B, \smol-1.7B, and \phims-1.5-1.3B.
Particularly, although \phims-1.5-1.3B has an advantage in ``textbook-like'' benchmarks such as \humanevalp, \mbppp, and \evoeval,
\ours-1.3B outperforms \phims-1.5-1.3B by 36\% on the more complex and practical \bigcodebench.
Also, \ours-1.3B beats \starcoderbase-3B, the predecessor of \starcodertwo-3B trained on 1T tokens, across all evaluated benchmarks.
Despite being trained on only 555B tokens, on \humaneval{+}, \ours-1.3B rivals and even surpasses models that have undergone significantly more extensive training, such as \starcodertwo-3B, \stablecode-3B, \codegemma-2B-v1.0, and \qwen{1.5}-1.8B. On \evoeval and \bigcodebench, \ours remains competitive.
Additionally, the table highlights the consistent improvement of \ours across its training phases: \oursone, \ourstwo, and the final \ours. Each phase builds on the previous one, with \ours achieving the highest scores in all benchmarks. This steady enhancement emphasizes the crucial role of high-quality and synthetic data in the final phase. Despite starting with the same data, each iteration of \ours narrows the gap with state-of-the-art models, demonstrating the efficacy of the overall training approach.

\subsection{Repo-level data in general pretraining}
\label{sec:ablation:repo}

In the general pretraining phase, we adopt \starcodertwo's approach to group file-level data randomly into repositories through a random concatenation of file contents~\cite{starcoder2}.
In \Cref{tab:ablation:repo}, we study two methods: (1) grouping files just by repository names, meaning that each training document can be a mix of multi-lingual code files if the repository is written in different languages, and (2) partitioning files into different programming languages before grouping them into repositories, meaning that each training document only focuses on one single language.

\begin{table*}[htbp]
\caption{Comparison of two methods for grouping repo-level data for pretraining. (1) ``Group by repo'' treats each repository as a single training unit with possibly mixed languages, and (2) ``Group by language and repo'' partitions data by programming language before grouping by repository.}

\label{tab:ablation:repo}
\centering
\begin{booktabs}{
    colspec={@{}lrrr@{}},
    % cell{1}{3}={c=2}{c},
    % cell{1}{1-2}={r=2}{m},
    % % cell{Y}{3-Z}={font=\bfseries}
    % row{X-Z} = {bg=lightb}
}
\toprule
Setting & \humaneval (+) & \mbpp (+) & \evoeval \\
\midrule
Group by repo & 12.8 (10.4) & 30.7 (25.9) & 7.0 \\ 
Group by language and repo & \textbf{17.1 (15.9)} & \textbf{33.9 (27.8)} & \textbf{7.4} \\
\bottomrule
\end{booktabs}%
\end{table*}

We can observe that the second approach, which we finally adopt in general pretraining, performs significantly better than the first one.

\subsection{Design choices in continued pretraining}
\label{sec:ablation:continue}

In continued pretraining, we source high-quality tokens from our pretraining corpus and train an improved base model. To obtain high-quality tokens, a model-based quality annotator is employed. In this section, we experiment with various design choices, including the training data for the annotator, the learning rate used in continued pretraining, and the optimal repetitions of high-quality tokens.

\paragraph{Model-based quality annotator}
Similar to \finewebedu~\cite{fineweb}, we train a linear head on top of the \snowarcm~\cite{arcticembed} embedding model to score each code file.
In \Cref{tab:ablation:annotator}, we experiment with 4 variants:
\begin{itemize}
    \item \annEdu: We prompt \mixtral-8x7B-Instruct~\cite{mixtral} to annotate the educational value of each code file (1 to 5). 400k annotations are used to train a linear regression head.
    For the following variants, similar to \dclm~\cite{dclm}, we sample negative documents randomly and change the positive parts only. A linear classification head is used instead.
    \item \annIns: Positives are a mix of 100k educational data (3.5+) bootstrapped from \annEdu and 100k high-quality instruction data from \magicoder~\cite{magicoder} and \scins~\cite{sc2instruct}.
    \item \annHq: Positives are 220k open-source, synthetic, high-quality code files~\cite{hqcode}.
    \item \annBest: Positives are a mix of 220k \annHq training data and 80k instruction data from \magicoder~\cite{magicoder} and \scins~\cite{sc2instruct}.
\end{itemize}
\begin{table*}[htbp]
\caption{Comparison of downstream performance by applying model-based quality annotators trained with different recipes to 10B continued pretraining.}
\label{tab:ablation:annotator}
\centering
\resizebox{\linewidth}{!}{%
\begin{booktabs}{
    colspec={@{}lQ[l,20em]rrr@{}},
    cell{2,3}{1}={c=2}{l}
    % cell{1}{3}={c=2}{c},
    % cell{1}{1-2}={r=2}{m},
    % % cell{Y}{3-Z}={font=\bfseries}
    % row{X-Z} = {bg=lightb}
}
\toprule
Annotator & Training data & \humaneval (+) & \mbpp (+) & \evoeval \\
\seprule
Pretrained model (no continued pretraining) & & 17.1 (15.9) & 33.9 (27.8) & 7.4\\
Continued pretraining on random 10B tokens & & 15.9 (12.8) & 30.7 (23.3) & 8.0\\
\midrule
\annEdu & 400k Mixtral annotations for educational scores (0--5) & 19.5 (16.5) & 27.8 (22.2) & 10.4\\
\annIns & 100k high \annEdu + 100k instruction data from \magicoder~\cite{magicoder} and \scins~\cite{sc2instruct} & 21.3 (18.3) & 37.3 (29.9) & 10.4 \\
\annHq & 220k open-source, synthetic high-quality code files~\cite{hqcode} & 19.5 (16.5) & 33.9 (26.7) & 9.2 \\
\annBest & 220k \annHq{} data mixed with 80k instruction data & \textbf{22.0 (18.3)} & \textbf{40.2 (33.1)} & \textbf{11.6} \\
\bottomrule
\end{booktabs}%
}
\end{table*}

After training the annotators, we first apply each annotator to the entire pretraining corpus to obtain a score for each file. Unlike \finewebedu, which only scans the top 2k characters, we scan the top, middle, and bottom parts of a code file and average the scores. We then rank the code files per language based on these scores and select the top percentile of documents until we reach approximately 10 billion tokens. We maintain the same mix ratio as used in pretraining. The table shows that \annBest, combining high-quality files and instruction data, achieves the best downstream performance.

\begin{figure}[htbp]
\centering
\includegraphics[width=0.6\linewidth]{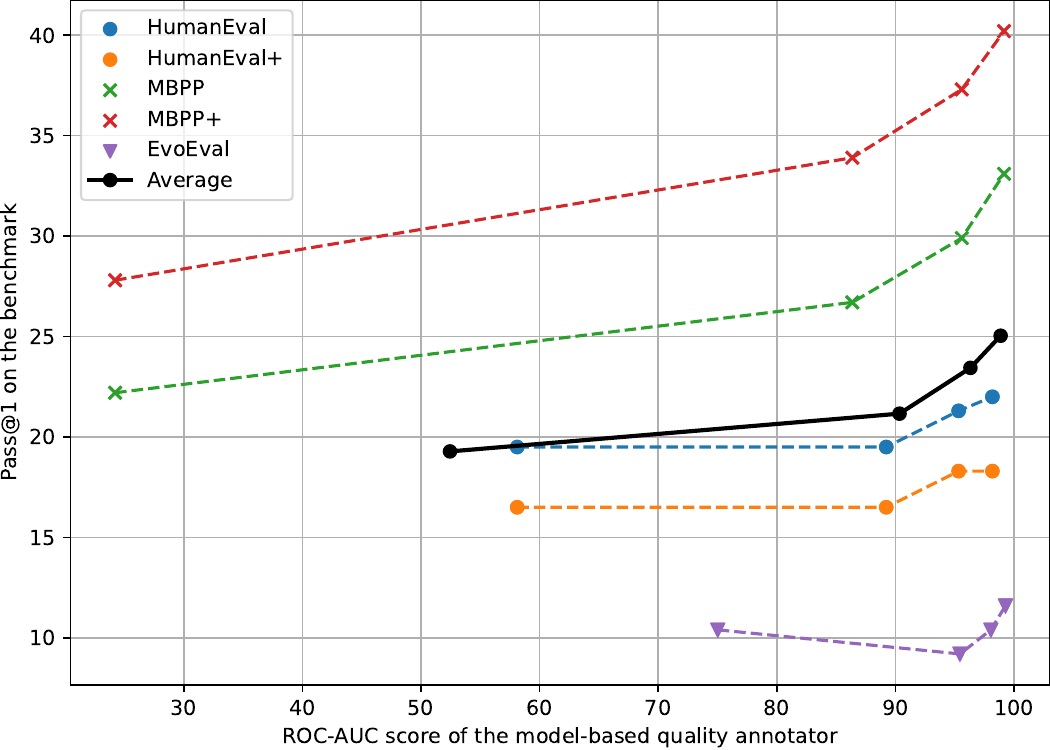}
\caption{Correlation between annotator ROC-AUC score and benchmark pass@1.}
\label{fig:ablation:correlation}
\end{figure}

We conduct an additional analysis in \Cref{fig:ablation:correlation}. For each annotator, we create a validation dataset with positives from code solution benchmarks and negatives from random pretraining data not seen during training. We use the ROC-AUC~\cite{rocauc} (Area Under the Receiver Operating Characteristic Curve) score to evaluate how well the annotator ranks benchmark data. The figure illustrates the correlation between per-benchmark ROC-AUC scores and benchmark pass rates. There is an almost consistent trend: higher ROC-AUC scores lead to better benchmark performance. A good ROC-AUC score indicates that the annotator effectively shapes the distribution of downstream tasks. Thus, the key to high-quality data is essentially the alignment with downstream application distributions.

\paragraph{Learning rate schedule}
We also study different learning rate schedules for continued pretraining in \Cref{tab:ablation:schedule}, including (1) a linear annealing starting from the minimum pretraining learning rate to zero, (2) a constant schedule using the minimum pretraining learning rate, and (3) a re-warmup to the maximum pretraining learning rate followed by a linear decay to zero.
\begin{table*}[htbp]
\caption{Comparison of different learning rate schedules in 10B continued pretraining using \annBest. Here $\mathMinLR = 5.3\times 10^{-5}$ and $\mathMaxLR = 5.3\times 10^{-4}$.}
\label{tab:ablation:schedule}
\centering
\begin{booktabs}{
    colspec={@{}llrrr@{}},
    % cell{1}{3}={c=2}{c},
    % cell{1}{1-2}={r=2}{m},
    % % cell{Y}{3-Z}={font=\bfseries}
    % row{X-Z} = {bg=lightb}
}
\toprule
Setting & Schedule & \humaneval (+) & \mbpp (+) & \evoeval \\
\seprule
Pretraining & $0 \to \mathMaxLR \to \mathMinLR$ & 17.1 (15.9) & 33.9 (27.8) & 7.4\\
\midrule
Linear & $\mathMinLR \to 0$ & 18.3 (16.5) & 37.0 (30.4) & 9.8 \\
Constant & $\mathMinLR \to \mathMinLR$ & 20.7 (18.3) & 39.4 (31.7) & 9.4 \\
Re-warmup & $0 \to \mathMaxLR \to 0$ & \textbf{22.0 (18.3)} & \textbf{40.2 (33.1)} & \textbf{11.6} \\
\bottomrule
\end{booktabs}%
\end{table*}

Empirically, we find that the re-warmup approach performs the best and use it consistently in all the other experiments with respect to continued pretraining.

\paragraph{Repetitions of high-quality data}
Finally, we scale up the token horizon from 10 billion to 50 billion in continued pretraining. One remaining question to address is determining the optimal repetitions for high-quality tokens. We experiment with repetitions ranging from 1 to 5, as shown in \Cref{tab:ablation:repeat}, by selecting the top percentile tokens ranked by \annBest. 
\begin{table*}[htbp]
\caption{Downstream performance with varying repetitions of high-quality data in 50B continued pretraining using \annBest.}

\label{tab:ablation:repeat}
\centering
\begin{booktabs}{
    colspec={@{}lrrr@{}},
    % cell{1}{3}={c=2}{c},
    % cell{1}{1-2}={r=2}{m},
    % % cell{Y}{3-Z}={font=\bfseries}
    % row{X-Z} = {bg=lightb}
}
\toprule
Repetition pattern & \humaneval (+) & \mbpp (+) & \evoeval \\
\seprule
Pretrained & 17.1 (15.9) & 33.9 (27.8) & 7.4\\
1 $\times$ 10.0B & 22.0 (18.3) & 40.2 (33.1) & 11.6 \\
\midrule
1 $\times$ 50.0B & 17.4 (14.0) & 41.5 (33.6) & 9.6 \\
2 $\times$ 25.0B & 23.2 (19.5) & 42.1 (34.7) & 9.2 \\
3 $\times$ 16.7B & 23.8 (18.9) & 42.3 (34.4) & 11.2 \\
4 $\times$ 12.5B & \textbf{26.2 (21.3)} & 40.2 (32.5) & \textbf{12.8} \\
5 $\times$ 10.0B & 20.1 (17.7) & \textbf{43.9 (36.0)} & 10.4 \\
\bottomrule
\end{booktabs}%
\end{table*}

In this context, the top percentile tokens are the highest quality tokens available. For example, 1 $\times$ 50B indicates one repetition of the top 50B tokens, while 4 $\times$ 12.5B denotes four repetitions of the top 12.5B tokens, ensuring that the selected tokens are of the best quality.
Based on the results in the table, repeating the high-quality tokens four times (4 $\times$ 12.5B) yields the best overall downstream performance across multiple evaluation metrics, showing the highest scores for \humaneval and \evoeval. Two repetitions (2 $\times$ 25.0B) and three repetitions (3 $\times$ 16.7B) also demonstrate strong performance, particularly in \texttt{mbpp}. Five repetitions (5 $\times$ 10.0B) achieve the highest \mbpp score but do not surpass the four repetitions in overall metrics. A single repetition (1 $\times$ 50.0B) shows the least improvement compared to multiple repetitions.

\section{Related Work}
\label{related}

\paragraph{Code pretraining corpus for language models}
Code data is essential to improving the reasoning capabilities of large language models (\llm{s})~\cite{tocodeornot,codecommensense,whichstagecode,llmcodewand,dscoderv2}. Typically, researchers obtain massive code pretraining data by crawling from public platforms hosting code repositories such as \github~\cite{li2023starcoder, codellama, dscoder, starcoder2, granite, dscoderv2}.
For example \stack v1~\cite{stackv1} is a 3.1 TB dataset consisting of permissively licensed source code mined from \github in 30 programming languages.
Its successor \stack v2~\cite{starcoder2}, built on the Software Heritage archive~\cite{softwareheritage}, is an order of magnitude larger, with a raw dataset of 67.5 TB spanning 619 programming languages.
However, directly using these massive unfiltered code for pretraining is suboptimal, because the code documents may contain undesired contents or duplicates. Therefore, further preprocessing steps are needed to downscale the raw corpus, which can include deduplication~\cite{li2023starcoder, codellama, dscoder, starcoder2, granite, dscoderv2, codegemma}, PII (Personally Identifiable Information) redaction~\cite{li2023starcoder, starcoder2, granite}, benchmark decontamination~\cite{li2023starcoder,starcoder2,dscoder,dscoderv2}, and model-based filtering~\cite{dscoderv2}.
As an example, \starcodertwo~\cite{starcoder2} selects only 3 TB of data for pretraining from the 67.5 TB total data available in \stack v2.
The code pretraining corpus of \ours follows a similar preprocessing pipeline, comprising approximately 400B unique tokens from a mix of filtered \stack v1 and \github crawls.
% \starcodertwo~\cite{starcoder2}
% \starcoder~\cite{li2023starcoder} and \starcodertwo~\cite{starcoder2} 

\paragraph{Model-based quality filtering}
In addition to common preprocessing steps like deduplication and heuristic filtering, a recent trend is using model-based quality filters to select high-quality pretraining data.
\phims-1~\cite{phi1} employs a random forest classifier trained on top of the CodeGen~\cite{nijkamp2023codegen} embedding layer on GPT-4 annotations, to assess the educational value of files. This filter selects high-quality \stack v1 and StackOverflow content, significantly enhancing coding performance.
\finewebedu~\cite{fineweb} employs a linear regressor built on \snowarcm~\cite{arcticembed}, an advanced embedding model based on BERT~\cite{bert}. This regressor, trained on 400k \llamathree~\cite{llama31} annotations rating the educational value (0-5) of \fineweb{} dataset documents, significantly enhances STEM performance.
\dclm-Baseline~\cite{dclm} uses a \fasttext~\cite{fasttext} filter trained on positives from OpenHermes~2.5~\cite{openhermes}, high-scoring posts from \texttt{r/ExplainLikeImFive}, and random negatives. It outperforms \finewebedu in top-10\% selection.
\deepseektwo~\cite{dscoderv2} follows \deepseekmath~\cite{deepseekmath} by leveraging a multi-stage \fasttext-based pipeline to recall high-quality code and math contents.
\llamathree~\cite{llama31} uses \fasttext for recognizing text referenced by Wikipedia~\cite{wikipedia} and Roberta-based~\cite{roberta} classifiers trained on \llamatwo~\cite{llama2} predictions.
While prior work focuses on initial pretraining, \ours demonstrates that high-quality data from the pretraining corpus can significantly enhance model performance during continued pretraining. We are also the first to uncover the secret of data quality, revealing the importance of matching data distribution with downstream tasks.

\paragraph{High-quality code data for pretraining}
\phims-1~\cite{phi1} is one of the first to study the impact of high-quality code data. It first uses a random forest classifier to filter out high-quality code data from \stack v1 and StackOverflow, and then creates synthetic textbook-like data and exercises using GPT-3.5~\cite{chatgpt}, showing significant coding performance with only 50B+ training tokens.
\deepseektwo~\cite{dscoderv2}, pretrained for around 14T tokens in total, achieves state-of-the-art coding performance, with a multi-stage \fasttext-based~\cite{fasttext} pipeline to recall web-related code data as well as high-quality \github code.
\ours utilizes a high-quality code annotator to extract high-quality code from pretraining datasets and generates synthetic files seeded from this high-quality data, adapting \magicoder \ossinstruct~\cite{magicoder} into pretraining.

\section{Conclusion}
We introduce \ours-1.3B, a high-performing code model that underscores the critical importance of data quality in the pretraining process. Trained on 555B tokens, \ours-1.3B achieves competitive results with state-of-the-art small code models while using significantly fewer tokens. Our three-stage pretraining process begins with 500B tokens of general pretraining on a raw code corpus, followed by 50B high-quality tokens scored by a quality annotator, and concludes with 5B tokens of synthetic data for further enhancement. This work demystifies the notion of high-quality data in code pretraining by demonstrating the key to high-quality data is its alignment with the distribution of downstream applications. Additionally, the paper offers practical guidelines for repo-level data grouping, learning rate scheduling, and the repetition of high-quality data, paving the way for more efficient and effective code model development.
% \section*{Limitations}

% \section*{{\textbf{Ethical and Social Implications}}}

\bibliography{custom}
\bibliographystyle{plain}

\appendix
\clearpage

\end{document}